\documentclass{article}
\usepackage{ijcai21}

\usepackage{times}
\usepackage{soul}
\usepackage{url}
\usepackage[hidelinks]{hyperref}
\usepackage[utf8]{inputenc}
\usepackage[small]{caption}
\usepackage{graphicx}
\usepackage{amsmath}
\usepackage{amsthm}
\usepackage{booktabs}
\usepackage{algorithm}
\usepackage{algorithmic}
\usepackage{epsfig}
\usepackage{graphicx}
\usepackage{amsmath}
\usepackage{amssymb}
\usepackage{color}
\usepackage{multirow}
\usepackage{subfigure}
\usepackage{underscore}

\title{Step-Wise Hierarchical Alignment Network for Image-Text Matching}

\author{
Zhong Ji
\and
Kexin Chen\and
Haoran Wang\thanks{indicates corresponding author}
\affiliations
  School of Electrical and Information Engineering, Tianjin University, Tianjin, China\\
\emails
\{jizhong, kxchen, haoranwang\}@tju.edu.cn
}

\begin{document}

\maketitle

\begin{abstract}
Image-text matching plays a central role in bridging the semantic gap between vision and language. The key point to achieve precise visual-semantic alignment lies in capturing the fine-grained cross-modal correspondence between image and text. Most previous methods rely on single-step reasoning to discover the visual-semantic interactions, which lacks the ability of exploiting the multi-level information to locate the hierarchical fine-grained relevance. Different from them, in this work, we propose a step-wise hierarchical alignment network (SHAN) that decomposes image-text matching into multi-step cross-modal reasoning process. Specifically, we first achieve local-to-local alignment at fragment level, following by performing global-to-local and global-to-global alignment at context level sequentially. This progressive alignment strategy supplies our model with more complementary and sufficient semantic clues to understand the hierarchical correlations between image and text. The experimental results on two benchmark datasets demonstrate the superiority of our proposed method.	
\end{abstract}

\section{Introduction}

\begin{figure}
	\begin{center}
		\includegraphics[height=6.6cm,width=0.98\linewidth]{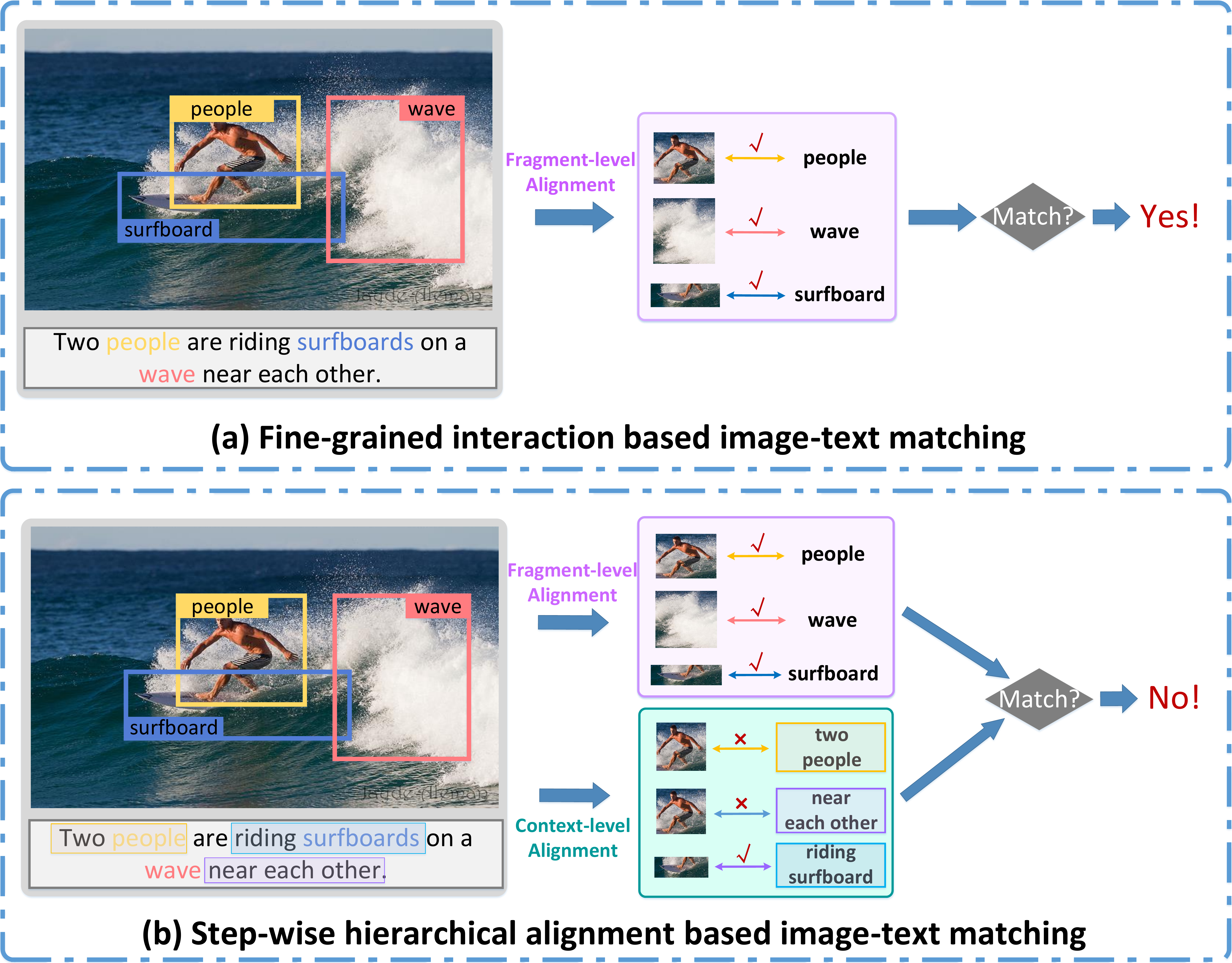}
	\end{center}
	\caption{Illustration of the hard negative example which may cause mismatching by current fine-grained interaction learning methods, while our proposed step-wise hierarchical alignment network can not only explore the fragment-level alignment, but also take the context-level information into consideration so as to discriminate the hard negative pairs more accurately. Note that the displayed image-text pair is mismatched.}
	\label{fig:fig1}
\end{figure}

Associating vision with language and exploring the correlations between them have attracted broad attention in the field of Artificial Intelligence during last decades. Thanks to the rapid development of deep learning \cite{krizhevsky2012imagenet}, plenty of areas have made tremendous progress in effectively bridging vision and language, such as visual question answering (VQA) \cite{antol2015vqa,anderson2018bottom}, image captioning \cite{zhang2019reconstruct,feng2019unsupervised}, image-text matching \cite{song2019polysemous,Wang2020CVSE}, visual grounding \cite{wang2020temporally}, and zero-shot learning \cite{yu2020episode}. In this paper, we focus on the task of image-text matching, which aims to provide flexible retrieval across samples from different modalities. It remains challenging since it is difficult for machines to understand the heterogeneous contents in images and texts so as to accurately infer the cross-modal similarity, especially when encountering hard negative example pairs which share similar semantics but convey slight differences (As shown in Fig.\ref{fig:fig1}).

To address this problem, early works \cite{frome2013devise,faghri2017vse++} usually map the whole images and texts into a shared embedding space, in which the cross-modal relevance can be directly measured by calculating the distance between heterogeneous samples. However, these global embedding based learning methods ignore that a visual or textual instance is actually composed of significant local fragments (\textit{e.g}., salient regions in image and key words in text) and insignificant ones. Therefore, represent them from a global view may not fully exploit the fine-grained details in fragment-level and introduce some noisy background information. Based on this observation, another stride of approaches \cite{karpathy2015deep,lee2018stacked} have been proposed to mitigate this issue by concentrating on the fine-grained interactions between image regions and textual words. These fine-grained interaction based learning methods align image and text by aggregating the local similarities of all available region-word pairs. Afterwards, recent works resort to attention mechanism \cite{vaswani2017attention,nam2017dual,lee2018stacked,hu2019multi} to pinpoint the discriminative parts of fragment-level representations, which can neatly align image and text via discovering more fine-grained cross-modal correspondence and further leads to promising matching performance.  

However, most existing fine-grained interaction learning based methods tend to discriminate image-text pairs according to the apparently distinct and distinguishable portions, which may consequently fail to distinguish those hard negative examples with similar semantic contents but slightly different context information. For example, as shown in Fig.\ref{fig:fig1}~(a), previous fine-grained interaction based methods may probably assign high matching score to the combination of the given image and the sentence ``Two people are riding surfboards on a wave near each other.'', due to they both contain the salient foreground objects, \textit{e.g.} ``\texttt{people}'', ``\texttt{surfboard}'' and ``\texttt{wave}''. The failure may be attributed to the inability of the previous methods, which only rely on single-step inference whilst pay less attention to the global context information that can provide some high-level clues between semantically similar samples. By contrast, when people want to precisely associate an image with a specific text, we typically not only attend to the fragment-level entities (``\texttt{people}'', ``\texttt{surfboard}'', ``\texttt{wave}''), but also leverage the context-level semantic information, \textit{e.g}., attributes (``\texttt{two}'') and relationships (``\texttt{near each other}''), as complementary hints to capture the complicated cross-modal correspondence. Accordingly, these observations motivate us to explore multi-step visual-semantic alignments for cross-modal correlation learning. 

To circumvent the aforementioned issues, in this work, we propose a \textbf{S}tep-wise \textbf{H}ierarchical \textbf{A}lignment \textbf{N}etwork (\textbf{SHAN}) for image-text matching, which performs hierarchical cross-modal alignment at two semantic levels, \textit{i.e.,} fragment-level alignment and context-level alignment. Specifically, we measure the relevance of an image-text pair by performing three steps of alignments progressively. First of all, we employ cross-attention mechanism to align image regions with textual words to achieve fragment-level local-to-local (L2L) alignment, followed by adopting an adaptive cross-modal feature fusion strategy to produce the visual and textual global context representations by fully exploiting the intermediate information produced by last alignment process. Then, based on the generated context representation, we perform global-to-local (G2L) alignment to  measure the cross-modal similarities at context-level. Finally, we further perform a global-to-global (G2G) cross-modal alignment by leveraging more context-level information. The main highlights of our proposed method are summarized as follows:		
		
\begin{enumerate}
	\item We propose a novel Step-wise Hierarchical Alignment Network to progressively explore the visual-semantic relevance for image-text matching, which leverages the multi-step hierarchical alignment to provide more complementary cues to capture the fine-grained cross-modal semantic correlation. 
	
	\item We conduct experiments on two public datasets: \textit{e.g.} Flickr30k and MS-COCO, and the quantitative experimental results validate that our model can achieve state-of-the-art performance on both datasets. Detailed visualization on L2L and G2L alignment step also qualitatively verify the effectiveness of our proposed method. 	
\end{enumerate}

\section{Related Work}

A stride of typical solutions for cross-modal image-text matching \cite{kiros2014unifying,ma2015multimodal,faghri2017vse++,wang2020stacked} can be classified as \textbf{global embedding based image-text matching}, which learns a joint embedding space for these heterogeneous data samples, so that the similarity of a given image-text pair can be directly compared. For example, \cite{kiros2014unifying} employed a convolution neural network (CNN) and an recurrent neural network (RNN) to extract the global representations for image and text respectively and utilized hinge-based triplet ranking loss to align these heterogeneous data. \cite{faghri2017vse++} further modified the ranking loss with hard negative mining and yielded encouraging cross-modal retrieval performance improvement. 

\begin{figure*}
	\begin{center}
		\includegraphics[height=5.8cm,width=0.98\linewidth]{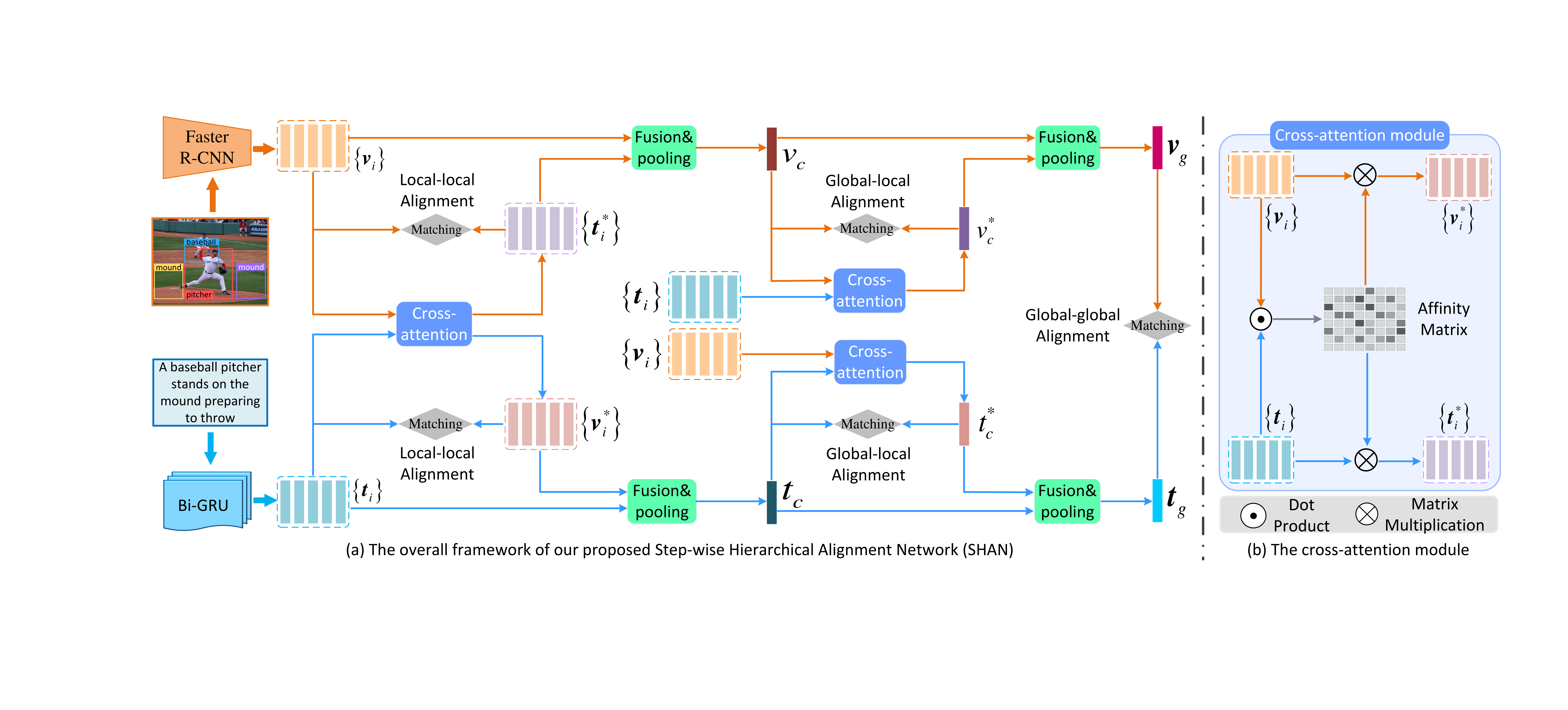}
	\end{center}
	\caption{The framework of our proposed Step-wise Hierarchical Alignment  Network (SHAN).}
	\label{fig:fig2}
\end{figure*}

Another line of studies \cite{lee2018stacked,hu2019multi,wang2019camp} can be summarized as \textbf{fine-grained interaction based image-text matching}, which focuses on exploring the fine-grained interactions between images and texts to obtain a deeper understanding of cross-modal correspondence. \cite{karpathy2015deep} developed a unified deep model to infer the latent alignment between textual segments and the implicit local regions of the image that they depict, and the global image-text similarity is finally calculated by aggregating all the region-word similarities. \cite{chen2020imram} proposed an iterative matching scheme to capture the correspondences between image regions and textual words with multiple steps of alignments, which is mostly relevant to our work. However, they paid less attention to explore the context-level information in images and texts. By contrast, we perform visual-semantic alignment from both perspectives including fragment-level and context-level.

\section{Our Method}

We construct a Step-wise Hierarchical Alignment Network (SHAN) to progressively explore the semantic alignments between image and text. The overall framework is illustrated in Fig.\ref{fig:fig2}~(a). Next, we will introduce the details of our proposed SHAN that contains following three parts: 1) Features representation for image and text; 2) The implementation details of designed Step-wise Hierarchical Alignment model and 3) Objective function for model learning.

\subsection{Feature Representation}

\subsubsection{Image Representation} 

For image feature extraction, we adopt an off-the-shelf object detection model (\textit{i.e.} Faster R-CNN) which is pre-trained on Visual Genome dataset to detect $k$ salient objects $\mathbf{O}{\kern 1pt} {\kern 1pt}  = {\kern 1pt} {\kern 1pt} \left\{ {{\mathbf{o}_1}, \cdot  \cdot  \cdot ,{\kern 1pt} {\kern 1pt} {\mathbf{o}_k}} \right\} \in {\mathbb{R} ^{k \times 2048}}$ in an image, then a fully-connect layer is followed to project them into $D$-dimension vectors:
\begin{equation}
{\mathbf{v}_i} = {\kern 1pt} {\kern 1pt} {\mathbf{W}_v}{\kern 1pt} {\mathbf{o}_i}{\kern 1pt} {\kern 1pt}  + {\kern 1pt} {\kern 1pt} {b_v}.
\end{equation}

In our experiments, the number of extracted salient regions for each image is set to 36.
 
\subsubsection{Text Representation} 

We extract the word-level features $\mathbf{W} = \left\{ {{\mathbf{w}_1}, \cdot  \cdot  \cdot {\kern 1pt} ,{\mathbf{w}_n}} \right\}$ to represent a given text, where $n$ is the number of words in the given sentence. Concretely, We first apply one-hot encoding on input words based on the whole vocabulary of the training dataset, then leverage a pre-trained word embedding model Glove and a learnable embedding layer to transform one-hot codes into fixed vectors and updatable vectors respectively. We then concatenate the two vectors to obtain the final word representations ${\mathbf{w}_j}$. 

Next, we employ a bidirectional GRU to enhance the word-level representations by capturing the context information in the sentence from both forward and backward directions:
\begin{equation}
\begin{array}{l}
\overrightarrow {{\mathbf{h}_j}}  = \overrightarrow {GRU} ({\mathbf{w}_j},\overrightarrow {{\mathbf{h}_{j - 1}}} ),\\
\overleftarrow {{\mathbf{h}_j}}  = \overleftarrow {GRU} ({\mathbf{w}_i},\overleftarrow {{\mathbf{h}_{j - 1}}} ),
\end{array}
\end{equation}
\MakeLowercase{where} $\overrightarrow {{\mathbf{h}_j}}$ and $\overleftarrow {{\mathbf{h}_j}}$ denote the GRU hidden states for forward and backward directions respectively, note that we set the hidden dimension of GRU as $D$, therefore finally we can get word features $\mathbf{T} = \left\{ {{\mathbf{t}_1}, \cdot  \cdot  \cdot {\kern 1pt} ,{\mathbf{t}_n}} \right\} \in {\mathbb{R} ^{n \times D}}$ by averaging $\overrightarrow {{\mathbf{h}_j}}$ and $\overleftarrow {{\mathbf{h}_j}}$: ${\mathbf{t}_j} = \frac{{\overrightarrow {{\mathbf{h}_j}}  + \overleftarrow {{\mathbf{h}_j}} }}{2}$.


\subsection{Step-wise Hierarchical Alignment}

\subsubsection{Fragment-level L2L Alignment}

In the first step, we carry out fine-grained fragment-level local-to-local matching between visual regions and textual words based on bidirectional cross-attention mechanism (see Fig.\ref{fig:fig2}~(b)). Mathematically, we first calculate the region-word affinity matrix as:
\begin{equation}
\mathbf{A} = ({{\mathbf{\tilde W}}_v}\mathbf{V}){({\mathbf{{\tilde W}}_t}\mathbf{T})^T},
\end{equation}
\MakeLowercase{where} $\mathbf{A} \in {\mathbb{R} ^{k \times n}}$ is the region-word affinity matrix and ${\mathbf{A}_{ij}}$ denotes the semantic similarity of the $i$-th region and the $j$-th word.

For region-queried cross-attention, each regional feature is used as a query to assign weights over the $n$ words, then we can construct an attended text-level representation $\mathbf{t}_i^*$ based on region feature $\mathbf{v}_i$ by applying weighted combination over the $n$ words, such process can be formulated as follows:
\begin{equation}
\mathbf{t}_i^* = \sum\limits_{j = 1}^n {{\alpha _{ij}}} {\mathbf{t}_j},{\kern 1pt} {\kern 1pt} {\kern 1pt} {\kern 1pt} {\alpha _{ij}}{\rm{ = }}\frac{{\exp (\lambda {\mathbf{A}_{ij}})}}{{\sum\nolimits_{j = 1}^n {\exp (\lambda {\mathbf{A}_{ij}})} }},
\end{equation}
\MakeLowercase{where} $\lambda$ is the temperature parameter of the softmax function. Similarly, we can obtain word attended image-level representations by regarding each word feature as query:
\begin{equation}
\mathbf{v}_j^* = \sum\limits_{i = 1}^k {{\beta _{ij}}} {\mathbf{v}_i},{\kern 1pt} {\kern 1pt} {\kern 1pt} {\kern 1pt} {\beta _{ij}}{\rm{ = }}\frac{{\exp (\lambda {\mathbf{A}_{ij}})}}{{\sum\nolimits_{i = 1}^k {\exp (\lambda {\mathbf{A}_{ij}})} }}.
\end{equation}

We then define the region-related L2L matching scores and word-related L2L matching scores as follows:
\begin{equation}
\begin{array}{l}
{R_v}({\mathbf{v}_i},\mathbf{t}_i^*) = \frac{{{\mathbf{v}_i}\mathbf{t}_i^{*T}}}{{\left\| {{\mathbf{v}_i}} \right\|\left\| {\mathbf{t}_i^*} \right\|}},i \in \left\{ {1, \cdot  \cdot  \cdot {\kern 1pt} ,k} \right\},\\
{R_t}({\mathbf{t}_j},\mathbf{v}_j^*) = \frac{{{\mathbf{t}_j}\mathbf{v}_j^{*T}}}{{\left\| {{\mathbf{t}_j}} \right\|\left\| {\mathbf{v}_j^*} \right\|}},j \in \left\{ {1, \cdot  \cdot  \cdot {\kern 1pt} ,n} \right\}.
\end{array}
\end{equation}

Finally, the L2L matching score for the target image $I$ and text $T$ is obtained through taking average of all region-based matching scores and word-based matching scores as follows:
\begin{equation}
\begin{aligned}
\begin{array}{l}
{S_{L2L}}(I,T) = {\mu _1}  {\sum\nolimits_{i = 1}^k {{R_v}({\mathbf{v}_i},\mathbf{t}_i^*)} } \\
\begin{array}{*{5}{c}}	
{\begin{array}{*{5}{c}}	
	{\begin{array}{*{5}{c}}		
		{}{}	
		\end{array}}&{}	
	\end{array}}&{}
\end{array}
 + (1 - {\mu _1}) {\sum\nolimits_{j = 1}^n {{R_t}({\mathbf{t}_j},\mathbf{v}_j^*)} },
\end{array}
\end{aligned}
\end{equation}
\MakeLowercase{where} ${\mu _1}$ is a hyper-parameter to balance the contributions of region-related L2L relevance and word-related L2L relevance.

\subsubsection{Context-level G2L Alignment} 

At the second step of our SHAN model, we implement a context-level global-to-local alignment mechanism to explore a deeper visual-semantic interaction. This alignment module plays a key role in exploiting the context-level information as complementary clues to understand the semantic correlations between image and text. The global representations for image and text in our model are obtained by implementing fusion and pooling operation on original object$/$word features and they are both enhanced by cross-modal attention. Specifically, we regard the region-attended text-level representations ${\mathbf{T}^*} = \left\{ {\mathbf{t}_1^*, \cdot  \cdot  \cdot {\kern 1pt} ,\mathbf{t}_k^*} \right\}$ as the aggregated message passed from textual to visual modality, and the word-attended image-level representations ${\mathbf{V}^*} = \left\{ {\mathbf{v}_1^*, \cdot  \cdot  \cdot {\kern 1pt} ,\mathbf{v}_n^*} \right\}$ as the aggregated message passed from visual to textual modality. Then we fuse the original fragment features with aggregated message from another modality to obtain the global context representation.

\begin{table*}[htbp]
	\setlength{\tabcolsep}{4mm}
	\begin{center}		
		\begin{tabular}{c|ccc|ccc|c}
			\hline 
			\multirow{2}{*}{Method}  & \multicolumn{3}{c|}{Text Retrieval}           & \multicolumn{3}{c|}{Image Retrieval}          & \multirow{2}{*}{Rsum} \\ \cline{2-7}
			& R@1           & R@5           & R@10          & R@1           & R@5           & R@10          &                      \\ \hline
			SCAN \cite{lee2018stacked}                     & 67.4          & 90.3          & 95.8          & 48.6          & 77.7          & 85.2          & 465.0                \\
			SAEM \cite{wu2019learning}                     & 69.1          & 91.0          & 95.1          & 52.4          & 81.1          & 88.1          & 477.0                \\
			RDAN \cite{hu2019multi}                     & 68.1          & 91.0          & 95.9          & 54.1          & 80.9          & 87.2          & 477.2                \\
			SGM \cite{wang2020cross}                      & 71.8          & 91.7          & 95.5          & 53.5          & 79.6          & 86.5          & 478.6                \\
			CAAN \cite{zhang2020context}                     & 70.1          & 91.6          & \textbf{97.2} & 52.8          & 79.0          & 87.9          & 478.6                \\  \hline
			SHAN-T2I                 & 72.5          & 92.3          & 95.8          & 53.6          & 78.6          & 85.5          & 478.3                \\
			SHAN-I2T                 & 70.6          & 91.7          & 95.5          & 50.5          & 77.1          & 85.2          & 470.6                \\
			SHAN-full                & \textbf{74.6} & \textbf{93.5} & 96.9 & \textbf{55.3} & \textbf{81.3} & \textbf{88.4} & \textbf{490.0}       \\ \hline 
		\end{tabular}
	\end{center}
	\caption{Experimental results on Flickr30k testing set in terms of Recall@$K$($R$@$K$).}
	\label{tab.1}
\end{table*}

\begin{table*}[htbp]
	\begin{center}		
		\setlength{\tabcolsep}{4mm}
		\begin{tabular}{c|ccc|ccc|c}
			\hline 
			\multirow{2}{*}{Method}  & \multicolumn{3}{c|}{Text Retrieval}           & \multicolumn{3}{c|}{Image Retrieval}          & \multirow{2}{*}{Rsum} \\ \cline{2-7}
			& R@1           & R@5           & R@10          & R@1           & R@5           & R@10          &                      \\ \hline
			SCAN \cite{lee2018stacked}                     & 72.7          & 94.8          & 98.4          & 58.8          & 88.4          & 94.8          & 507.9                \\
			SAEM \cite{wu2019learning}                     & 71.2          & 94.1          & 97.7          & 57.8          & 88.6          & 94.9          & 504.3                \\
			RDAN \cite{hu2019multi}                     & 74.6          & 96.2          & \textbf{98.7} & 61.6          & 89.2          & 94.7          & 515.0                \\
			SGM \cite{wang2020cross}                      & 73.4          & 93.8          & 97.8          & 57.5          & 87.3          & 94.3          & 504.1                \\
			CAAN \cite{zhang2020context}                     & 75.5          & 95.4          & 98.5          & 61.3          & \textbf{89.7} & 95.2          & 515.6                \\ \hline
			SHAN-T2I                 & 75.9          & 96.1          & \textbf{98.7}          & 60.7          & 88.2          & 94.2          & 513.8                \\
			SHAN-I2T                 & 73.0          & 95.8          & 97.9          & 58.5          & 87.3          & 94.0          & 506.5                   \\
			SHAN-full                & \textbf{76.8} & \textbf{96.3} & \textbf{98.7} & \textbf{62.6} & 89.6 & \textbf{95.8} & \textbf{519.8}       \\ \hline 
		\end{tabular}	
	\end{center}
	\caption{Experimental results on MS-COCO 1$K$ testing set in terms of Recall@$K$($R$@$K$).}
	\label{tab.2}
\end{table*}

Taking the visual regional features $\mathbf{V} = \left\{ {{\mathbf{v}_1}, \cdot  \cdot  \cdot {\kern 1pt} ,{\mathbf{v}_k}} \right\}$ and region-attended text-level representations ${\mathbf{T}^*} = \left\{ {\mathbf{t}_1^*, \cdot  \cdot  \cdot {\kern 1pt} ,\mathbf{t}_k^*} \right\}$ as an example. We first perform element-wise fusion operation in a dynamic way as follows:
\begin{equation}
\begin{array}{l}
\mathbf{v}_i^c = g * {\mathbf{v}_i} + (1 - g) * \mathbf{t}_i^*,\\
g = sigmoid({\mathbf{W}_g}cat({\mathbf{v}_i},\mathbf{t}_i^*) + {b_g}),
\end{array}
\end{equation}
\MakeLowercase{where} $\mathbf{v}_i^c$ denotes the contextual information provided by the i-th region ${\mathbf{v}_i}$ and its attended semantic knowledge $\mathbf{t}_i^*$ from textual modality, $g \in [0,1]$ is a gate value to adaptively balance the importance of fused information from $\mathbf{v}_i$ and $\mathbf{t}_i^*$, $cat()$ denotes concatenation on $\mathbf{v}_i$ and $\mathbf{t}_i^*$. Empirically, we found that such an adaptive fusion strategy outperforms the traditional average pooling operation.

For the $k$ regions in an image, we can acquire $k$ fused visual contextual representations ${\mathbf{V}^c} = \left\{ {\mathbf{v}_1^c, \cdot  \cdot  \cdot {\kern 1pt} ,\mathbf{v}_k^c} \right\} \in {\mathbb{R} ^{k \times D}}$. Similarly, for the $n$ words in a sentence, $n$ fused textual contextual representations ${\mathbf{T}^c} = \left\{ {\mathbf{t}_1^c, \cdot  \cdot  \cdot {\kern 1pt} ,\mathbf{t}_n^c} \right\} \in {\mathbb{R} ^{n \times D}}$ are also obtained. Then we use a simple self-attention mechanism to aggregate these separate features of $k$ regions and $n$ words into one global vector to represent the whole image and the whole sentence: 
\begin{equation}
\begin{array}{l}
{\gamma _v} = softmax({\mathbf{{\hat W}}_v}{\mathbf{V}^{cT}}),{\kern 1pt} {\kern 1pt} {\kern 1pt} {\kern 1pt} {\kern 1pt} {\mathbf{v}_c} = {\gamma _v}{\mathbf{V}^c},\\
{\gamma _t} = softmax({\mathbf{{\hat W}}_t}{\mathbf{T}^{cT}}),{\kern 1pt} {\kern 1pt} {\kern 1pt} {\kern 1pt} {\kern 1pt} {\kern 1pt} {\kern 1pt} {\mathbf{t}_c} = {\gamma _t}{\mathbf{T}^c},
\end{array}
\end{equation}
\MakeLowercase{where} ${\gamma _v} \in {\mathbb{R} ^{1 \times k}}$ and ${\gamma _t} \in {\mathbb{R} ^{1 \times n}}$ denote the weights for convex combination generated by two linear projection layers respectively, ${\mathbf{v}_c} \in {\mathbb{R} ^{1 \times D}}$ is the global context representation for image $I$, and ${\mathbf{t}_c} \in {\mathbb{R} ^{1 \times D}}$ is the global context representation for text $T$.

Based on the generated visual and textual global context representations, similar to L2L alignment module, we then perform G2L alignment by resorting to bidirectional cross-attention mechanism. Specifically, for visual global-queried cross-modal attention, $\mathbf{v}_c$ is used as an anchor vector to assign weights over $n$ different words according to the semantic similarity between $\mathbf{v}_c$ and $n$ word features, then we use a convex combination to aggregate textual word features into a visual global-attended text-level representation:
\begin{equation}
{\alpha _j} = \frac{{\exp (sim({\mathbf{v}_c},{\mathbf{t}_j}))}}{{\sum\nolimits_{j = 1}^n {\exp (sim({\mathbf{v}_c},{\mathbf{t}_j}))} }},{\kern 1pt} {\kern 1pt} {\kern 1pt} {\kern 1pt} {\kern 1pt} \mathbf{t}_c^* = \sum\limits_{j = 1}^n {{\alpha _j}} {\mathbf{t}_j},
\end{equation}
\MakeLowercase{where} $sim(.,.)$ denotes the distance function to measure the similarity of the two vectors, and $sim()$ represents the Cosine distance function. Similarly, for textual global-queried cross-attention, we can obtain the textual global-attended image-level representation as follows:
\begin{equation}
{\beta _i} = \frac{{\exp (sim({\mathbf{t}_c},{\mathbf{v}_i}))}}{{\sum\nolimits_{i = 1}^k {\exp (sim({\mathbf{t}_c},{\mathbf{v}_i}))} }},{\kern 1pt} {\kern 1pt} {\kern 1pt} {\kern 1pt} {\kern 1pt} \mathbf{v}_c^* = \sum\limits_{i = 1}^k {{\beta _i}} {\mathbf{v}_i}.
\end{equation}

Then the visual global-related G2L matching score and textual global-related G2L matching score are derived as follows:
\begin{equation}
{R_v}({\mathbf{v}_c},\mathbf{t}_c^*) = \frac{{{\mathbf{v}_c}\mathbf{t}_c^{*T}}}{{\left\| {{\mathbf{v}_c}} \right\|\left\| {\mathbf{t}_c^*} \right\|}},{\kern 1pt} {\kern 1pt} {\kern 1pt} {\kern 1pt} {\kern 1pt} {R_t}({\mathbf{t}_c},\mathbf{v}_c^*) = \frac{{{\mathbf{t}_c}\mathbf{v}_c^{*T}}}{{\left\| {{\mathbf{t}_c}} \right\|\left\| {\mathbf{v}_c^*} \right\|}}.
\end{equation}

Afterwards, we define the final G2L matching score between image $I$ and text $T$ by summing up the two matching scores:
\begin{equation}
{S_{G2L}}(I,T) = {\mu _2} \times {R_v}({\mathbf{v}_c},\mathbf{t}_c^*) + (1 - {\mu _2}) \times {R_t}({\mathbf{t}_c},\mathbf{v}_c^*),
\end{equation}
\MakeLowercase{where} ${\mu _2}$ is a hyper-parameter to balance the contributions of the two matching scores.

\subsubsection{Context-level G2G Alignment} 

On the top level of our SHAN model, we aim to further align the image and text from a global perspective. Specifically, the context-level global representations for image and text are obtained via fusing the intermediate semantic information produced during G2L alignment module, \textit{i.e.} fusing ${\mathbf{v}_c}$ and $\mathbf{t}_c^*$ to obtain the final visual context-level global representation , and fusing ${\mathbf{t}_c}$ and $\mathbf{v}_c^*$ to obtain the final textual context-level global representation:
\begin{equation}
\begin{array}{l}
{\mathbf{v}_g} = {\mathbf{v}_c} \oplus \mathbf{t}_c^*,\\
{\mathbf{t}_g} = {\mathbf{t}_c} \oplus \mathbf{v}_c^*,
\end{array}
\end{equation}
\MakeLowercase{where} $ \oplus $ denotes element-wise sum. Our hypothesis is that after L2L and G2L alignment, the generated representations ${\mathbf{v}_c}$ and ${\mathbf{t}_c}$ have already captured rich implicit information regarding fine-grained visual-textual correlations, such a cross-attention based fusion strategy would further encourage the mutual information propagation between both modalities so as to learn more discriminative context-level features.

Then the overall G2G matching score between image $I$ and text $T$ is derived by computing the cosine distance of ${\mathbf{v}_g}$ and ${\mathbf{t}_g}$:
\begin{equation}
{S_{G2G}}(I,T) = \frac{{{\mathbf{v}_g}\mathbf{t}_g^T}}{{\left\| {{\mathbf{v}_g}} \right\|\left\| {{\mathbf{t}_g}} \right\|}}
\end{equation}

\subsection{Learning Objective}
As aforementioned, our model eventually acquire three matching scores (\textit{e.g.}${S_{L2L}}(I,T)$, ${S_{G2L}}(I,T)$ and ${S_{G2G}}(I,T)$) at different levels for a given image-text pair $<I, T>$, and the final cross-modal similarity can be defined as follows:
\begin{equation}
S(I,T) = {S_{L2L}}(I,T) + {S_{G2L}}(I,T) + {S_{G2G}}(I,T).
\end{equation}

Then, we train our model with a bidirectional hinge-based triplet ranking loss  \cite{faghri2017vse++,Wang2020CVSE} that enforces the similarity of matched image-text pair to be higher than that of unmatched ones.
\begin{equation}
	\begin{array}{l}
	L(I,T) = \max (0,m - S(I,T) + S(I,\hat T))\\
	{\kern 1pt} {\kern 1pt} {\kern 1pt} {\kern 1pt} {\kern 1pt} {\kern 1pt} {\kern 1pt} {\kern 1pt} {\kern 1pt} {\kern 1pt} {\kern 1pt} {\kern 1pt} {\kern 1pt} {\kern 1pt} {\kern 1pt} {\kern 1pt} {\kern 1pt} {\kern 1pt} {\kern 1pt} {\kern 1pt} {\kern 1pt} {\kern 1pt} {\kern 1pt} {\kern 1pt} {\kern 1pt} {\kern 1pt} {\kern 1pt} {\kern 1pt} {\kern 1pt} {\kern 1pt} {\kern 1pt} {\kern 1pt} {\kern 1pt} {\kern 1pt} {\kern 1pt} {\kern 1pt} {\kern 1pt}  + \max (0,m - S(I,T) + S(\hat I,T)),
	\end{array}
\end{equation}
\MakeLowercase{where} $m$ is a tuning margin parameter. $\hat T = argma{x_{t \ne T}}S(I,t)$ and $\hat I = argma{x_{i \ne I}}S(i,T)$ stand for contrastive negative samples in a mini-batch. The entire model is trained in an end-to-end manner.

\section{Experiments}

\subsection{Dataset and Evaluation Metric}

\subsubsection{Dataset} 

Two benchmark datasets are used in our experiments to testify the performance of our method: (1) \textbf{Flickr30k} contains 31783 images and each image is annotated with 5 sentences. Following \cite{karpathy2015deep}, we split the dataset into 1000 test images, 1000 validation images and 29000 training images. (2) \textbf{MS-COCO} is another large-scale image captioning dataset with 123287 images and each image is relative with 5 descriptions. We follow  \cite{lee2018stacked} to split the dataset into 5000 images for validation, 5000 images for testing and the rest 113287 images for training. We report the matching performance by averaging the 5-folds over $1k$ test images.

\begin{figure*}
	\begin{center}
		\includegraphics[height=8.6cm,width=0.74\linewidth]{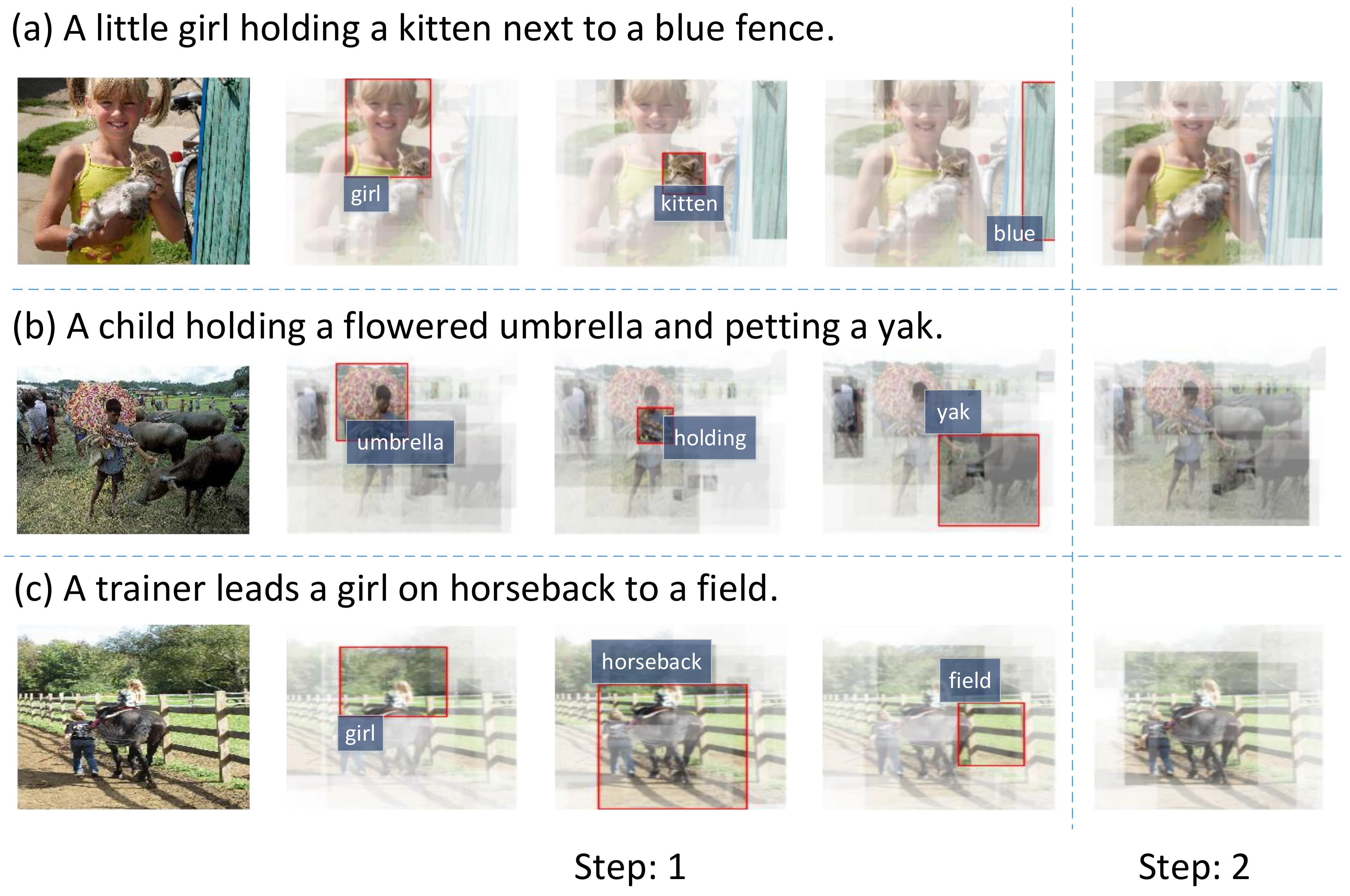}
	\end{center}
	\caption{Visualization of attention weights in L2L and G2L alignment step. Specifically, at the first alignment step, each word feature is used as a query to assign weights on salient image regions, and at the second alignment step, textual context-level  global features are leveraged as queries to assign different weights on salient image regions. Based on the generated cross-attention weights, we visualize the corresponding attended visual information produced by L2L and G2L alignment step progressively.}
	\label{fig:fig3}
\end{figure*}

\subsubsection{Evaluation Metric} 

We take Recall@$K$ ($R$@$K$, $K$ = 1, 5, 10) that describes the fraction of ground truth instance being retrieved at the top 1, 5, 10 results as the evaluation metric. Following previous works, we evaluate our approach on both text retrieval (find the most related text given an image) and image retrieval (find the matched image described by certain text) tasks. Additionally, the overall matching metric “$Rsum$” that sums over all recall values is also used to compare our model with other state-of-the-art methods.

\subsection{Implementation Details}

We present three variants of SHAN model to systematically investigate the effectiveness of our proposed approach: (1) SHAN-T2I applies hierarchical alignment only in text-to-image direction, the global representation for image in G2G alignment module is obtained simply by taking average over extracted regional features. (2) SHAN-I2T applies hierarchical alignment only in image-to-text direction, the global representation for text in G2G alignment module is obtained by taking average over word features. (3) SHAN-full applies hierarchical alignment both in text-to-image and image-to-text direction. We train our model with Adam optimizer for 30 epochs on each dataset. The dimension of joint embedding space for image regions and textual words are set to 1024, and the dimension of word embeddings is set to be 300, other parameters are empirically set as follows: ${\mu _1}{\rm{ = 0}}{\rm{.3}}$, ${\mu _{\rm{2}}}{\rm{ = 0}}{\rm{.5}}$, $\lambda {\rm{ = 15}}$ and $m = 0.2$.

\subsection{Comparison with State-of-the-art Methods}
Table \ref{tab.1} and Table \ref{tab.2} display the performance comparison between our SHAN model and several state-of-the-art methods on Flickr30k and MS-COCO dataset, respectively. We can observe that our model achieves the best retrieval results on both datasets. Specifically, for text retrieval, our SHAN-full achieves a maximal improvement of 2.8\% and 1.3\% in terms of R@1 score on Flickr30k and MS-COCO, respectively. As for image retrieval, our best model reaches at 55.3\% R@1 score and a 62.6\% R@1 score on Flickr30k and MS-COCO respectively, which also surpasses other state-of-the-art methods. Similar performance improvement can be seen from other evaluation metrics (\textit{e.g.} R@5, R@10).

\begin{table}[htbp]
	\begin{center}
		\setlength{\tabcolsep}{1.3mm}
		\resizebox{0.49\textwidth}{1.55cm}{
		\begin{tabular}{c|cc|cc|c}
			\hline 
			\multirow{2}{*}{Method}                            
			& \multicolumn{2}{c|}{Text Retrieval}     & \multicolumn{2}{c|}{Image Retrieval}  & \multirow{2}{*}{Rsum}    \\ \cline{2-5}  
			& R@1		& R@5			& R@1		& R@5			&		 \\ \hline
			SHAN-T2I $_{Glove}$	            & 74.8      & 94.9     		& 59.9 		& 88.4 			& 510.9		 \\
			SHAN-T2I $_{\Delta Glove}$	    & 74.9      & 95.2     	    & 59.7 		& 88.0 			& 510.7		 \\
			SHAN-T2I $_{random}$	        & 75.0      & 96.0     		& 59.6 		& 87.6 			& 510.2		 \\
			SHAN-T2I $_{L2L}$	            & 72.4      & 94.5     		& 58.2 		& 87.5 			& 504.4		 \\
			SHAN-T2I $_{L2L+G2L}$	        & 74.9      & 95.5     		& 59.3 		& 88.4 			& 510.5		 \\ \hline
			SHAN-T2I 	                    & \textbf{75.9}      & \textbf{96.1}    	    & \textbf{60.7} 		& \textbf{88.2} 			& \textbf{513.8}	 \\
			\hline 
		\end{tabular}
		}
		\caption{Ablation experimental results on MS-COCO $1k$ testing set in terms of Recall@$K$($R$@$K$).}
		\label{tab.3}
	\end{center}
\end{table}


\subsection{Ablation Studies}

In this section, we perform ablation studies to investigate the contributions of each component in our proposed SHAN approach. We first validate the effect of different word embedding models by training SHAN-T2I with fixed Glove embedding only (SHAN-T2I$_{\text{Glove}}$), fine-tuned Glove embedding only (SHAN-T2I$_{\text{$\Delta$ Glove}}$) and a learnable embedding layer only (SHAN-T2I$_{\text{random}}$). Then, we train SHAN-T2I by removing the G2L alignment module (SHAN-T2I$_{\text{L2L}}$) and G2G alignment module (SHAN-T2I$_{\text{L2L + G2L}}$) respectively to testify the impact of the proposed hierarchical alignment framework. The experimental results are shown in Table \ref{tab.3}. From Table \ref{tab.3}, we can draw the following conclusions: (1) The combination of pre-trained Glove embedding model with learnable embedding model can lead to better performance, in comparison to using Glove only or learnable embedding model only; (2) The proposed context-level global-to-local alignment module and context-level global-to-global alignment module are both validated to be effective by providing considering performance boost.

\subsection{Visualization}

From the qualitative results in Fig.\ref{fig:fig3}, we can observe that the words are appropriately matched with its corresponding depicted regions at alignment step 1. For example, in Fig. \ref{fig:fig3}~(b), our model can well generate interpretable attention weights on nouns like ``\texttt{umbrella}'' and ``\texttt{yak}'', as well as the actions like ``\texttt{holding}''. It reveals that L2L alignment module is really conducive to bridging the gap between heterogeneous fragments. At step 2 of alignment, we can see that the meaningful image regions contributing to understanding the global context information are highlighted and trivial ones are suppressed. Overall, these observations confirm the effectiveness of our proposed alignment modules. 	

\section{Conclusion}  

In this paper, we proposed a Step-wise Hierarchical Alignment Network (SHAN) for image-text matching. Specifically, fragment-level local-to-local alignment, context-level global-to-local alignment and context-level global-to-global alignment were performed progressively to learn the visual-semantic correspondence between instances from different modalities. We conducted extensive experiments and ablation studies on two widely-used benchmark datasets, and the experimental results demonstrated our method can achieve competitive results.

\bibliographystyle{named}
\bibliography{ijcai21}

\end{document}